\documentclass[conference]{IEEEtran}
\IEEEoverridecommandlockouts
\usepackage{cite}
\usepackage{amsmath,amssymb,amsfonts}
\usepackage{algorithmic}
\usepackage{graphicx}
\usepackage{textcomp}
\usepackage{xcolor}
\usepackage[ruled,vlined,linesnumbered]{algorithm2e}
\SetKwComment{Comment}{$\triangleright$\ }{}

\usepackage{tabularx}
\usepackage{booktabs}
\usepackage{multirow}

\usepackage{hyperref}
\hypersetup{hidelinks}

\newlength{\widestlhs}
\newcommand{\assign}[1]{\makebox[\widestlhs][l]{\ensuremath{#1}}\ \ensuremath{\gets}\ }

\def\BibTeX{{\rm B\kern-.05em{\sc i\kern-.025em b}\kern-.08em
    T\kern-.1667em\lower.7ex\hbox{E}\kern-.125emX}}
\begin{document}

\title{Complementary Attention Head Pruning for Efficient Transformers\\
\thanks{{This research was supported by the Israeli Ministry of Innovation, Science, and Technology (Grant No. 0008556).}}
\thanks{\copyright~2026 IEEE. Personal use of this material is permitted. Permission from IEEE must be obtained for all other uses, in any current or future media, including reprinting/republishing this material for advertising or promotional purposes, creating new collective works, for resale or redistribution to servers or lists, or reuse of any copyrighted component of this work in other works.}
}

\author{\IEEEauthorblockN{Yaniv Livertovsky}
\IEEEauthorblockA{\textit{Bar-Ilan University}\\
Ramat Gan, Israel \\
yaniv.livertovsky@biu.ac.il}
\and
\IEEEauthorblockN{Shahar Somin}
\IEEEauthorblockA{\textit{Bar-Ilan University}\\
Ramat Gan, Israel \\
shahar.somin@biu.ac.il}
\and
\IEEEauthorblockN{Gonen Singer}
\IEEEauthorblockA{\textit{Bar-Ilan University}\\
Ramat Gan, Israel \\
gonen.singer@biu.ac.il}
}
\maketitle

\begin{abstract}
The remarkable success of Transformer-based models in natural language processing stems from architectural scaling, which leads to a large number of parameters and hinders deployment in resource-constrained environments.
While structured pruning offers a pathway to compression, existing state-of-the-art methods often rely on gradient-based importance ranking or stochastic gating, which suffer from instability, structural degeneration, and the need for extensive manual hyperparameter tuning. 
In this paper, we introduce CAHP (Complementary Attention Head Pruning), a novel post-hoc framework that redefines head selection as a global graph-theoretical problem. 
Rather than evaluating heads in isolation, CAHP utilizes graph-based clustering combined with information-theoretic distance measures to identify and preserve a topologically diverse subset of complementary attention heads. 
Without requiring a predefined sparsity level or pruning ratio, the framework automatically determines the number of selected attention heads across layers by identifying a diminishing marginal performance curve, where pruning additional heads leads to a sharp degradation in performance, as determined by the chosen polynomial degree. 
Extensive evaluations on the SST-5 and MNLI benchmarks, {across different Transformer model scales,} demonstrate that CAHP consistently outperforms competitive baselines, particularly in high-compression regimes. 
Furthermore, our structural analysis shows that CAHP avoids the “proximity bias” of gradient-based pruning methods, which tend to preserve heads mainly in layers close to the output, and instead retains a functionally critical set of attention heads in the model’s intermediate layers.
\end{abstract}


\section{Introduction}
The rapid evolution of neural networks has led to their dominance across diverse fields, ranging from numerical regression using linear models \cite{hornik1989multilayer} and computer vision via convolutional networks \cite{he2016deep}, to complex natural language processing tasks using Transformer-based architectures \cite{vaswani2017attention}. 
However, the remarkable performance of these models comes at the cost of substantial over-parameterization. With parameter counts reaching into the hundreds of millions, deploying these ``heavyweight'' models on resource-constrained hardware remains a major practical hurdle.

While model compression via structured pruning has emerged as a viable solution \cite{anwar2017structured, levin2025automatic}, targeting entire components such as neurons, channels, or attention heads, a gap exists between theoretical ``prunability'' and practical implementation. Prior studies show that a fraction of attention heads are redundant and removable with modest accuracy loss, though sensitivity varies by task and layer \cite{voita2019analyzing, michel2019sixteen}. Yet, for the end-user, these methods remain incomplete. They often require manual intervention to determine the pruning ratio to identify the balance between model size and performance. This manual tuning regime effectively prevents pruning from being a standard part of the deployment pipeline in environments where time and computational resources are limited.


{To address these limitations, we present Complementary Attention Head Pruning (CAHP), an automated post-hoc pruning framework for Transformer models that optimizes head allocation based on diminishing marginal performance gains rather than predefined sparsity targets. Our approach builds upon the graph-based complementary selection and knee-point detection principles of Automatic Complementary Separation Pruning (ACSP) \cite{levin2025automatic}, extending its layer-wise logic from convolutional and linear settings into a global selection framework tailored for self-attention architectures. By treating the set of attention heads as a unified graph space, CAHP enables cross-layer redistribution, preserving the model's functional core regardless of depth. To adapt to the Transformer domain, we introduce attention-based signatures for behavioral profiling and padding-aware interpolation to normalize variable-length sequences. Furthermore, a gradient-based salience proxy that estimates head importance replaces static weights with a dynamic sensitivity measure. These adaptations navigate the structural dependencies of language models while maintaining the theoretical advantages of graph-based selection.}

By automatically identifying a diminishing marginal performance curve, the framework determines the number and layer-wise distribution of attention heads without requiring a predefined sparsity level or pruning ratio. The selection point is defined by the chosen polynomial degree, beyond which pruning additional heads leads to a sharp performance degradation, thereby eliminating the need for manual, trial-and-error ratio tuning. Unlike previous approaches that require extensive hyperparameter searches or simultaneous training-and-pruning procedures \cite{li2021differentiable, ding2024pruning, gale2019state}, our framework allows a user to apply this selection to a pre-trained model and obtain a refined architecture that recovers near-baseline accuracy through a resource-efficient, single-stage fine-tuning process.

The main contributions of this work are summarized below:
\begin{itemize}

    \item {\textbf{Complexity-Driven Global Structured Pruning for Transformers:} We introduce a post-hoc framework evolving beyond layer-wise pruning, treating the entire model as a unified graph space. By utilizing a complexity-driven selection mechanism based on diminishing marginal performance gains, CAHP automatically determines the optimal cross-layer distribution of attention heads, eliminating the need for predefined sparsity targets or manual hyperparameter tuning.}


    \item {\textbf{Complementary-Based Head Selection:} Moving beyond standard importance-based metrics and gradient-centric methods, we introduce a complementarity-driven selection strategy based on the principle of complementary separation. By utilizing padding-aware statistical signatures and a gradient-based salience proxy, our framework identifies and retains a functionally diverse subset of attention heads providing nonredundant information, ensuring the pruned architecture preserves the functional capacity and representational diversity of the model.}


    \item {\textbf{Comprehensive End-to-End Pipeline:} We present a simple post-hoc pipeline that converts standard pretrained Transformers into pruned, deployment-ready models. By combining automatic head selection with a lightweight finetuning step, our framework provides a stable and principled solution for reducing model size, ensuring superior structural consistency across multiple runs without the need for manual effort or hyperparameter tuning.}
\end{itemize}

\smallskip
The code for this work and any supporting materials are publicly available at \url{https://github.com/yanivlivert/cahp}.

\section{Related Work}

\subsection{Structured Pruning and Head Redundancy}
Model pruning aims to reduce the memory and computational footprint of deep neural networks by removing redundant parameters \cite{levin2025automatic, han2015learning}. While early work focused on unstructured, magnitude-based weight pruning \cite{han2015learning}, modern deployment typically requires structured pruning to achieve hardware-level acceleration \cite{anwar2017structured, levin2025automatic}. Structured pruning removes entire architectural blocks, such as convolutional filters or attention heads, maintaining the dense matrix formats optimized for standard processors.

In the context of Transformer architectures, diagnostic studies have revealed significant over-parameterization in multi-head attention (MHA) mechanisms. Prior work \cite{michel2019sixteen} has shown that many attention heads can be ablated individually or in large groups without substantial loss in accuracy, often finding that a single head per layer is sufficient for model performance. Additional analyses have demonstrated that attention heads often assume specialized functional roles, such as positional, syntactic, or rare-word processing, while indicating that a large fraction of {nonspecialized} heads can be safely removed without degrading overall performance \cite{voita2019analyzing}. Despite these insights, early attempts at head pruning remained largely post-hoc and manual, relying on simple ablation or sensitivity heuristics \cite{michel2019sixteen}. Such strategies evaluate heads in isolation and lack a mechanism to account for functional overlap between them, leaving a critical gap for automated methods that can identify a diverse, complementary subset of components without exhaustive human intervention.

\subsection{Optimization-Based and Attribution Methods}
To systematize attention head pruning, several works have proposed differentiable optimization-based approaches that integrate pruning decisions into the training objective. Differentiable Subset Pruning (DSP) \cite{li2021differentiable} employs Gumbel-Softmax relaxations to learn discrete head masks, while Pruning with Almost-Sure Sparsity (PASS) \cite{ding2024pruning} utilizes hard concrete distributions to enforce sparsity constraints. These methods jointly optimize model parameters and pruning masks through soft-gating rather than physically removing attention heads.

While effective, such optimization-based approaches typically require manually specified global sparsity targets, defined either as a fixed number of retained heads or as a sparsity ratio. In addition, they are sensitive to task-specific hyperparameters, such as temperature schedules or regularization coefficients, and often involve computationally expensive retraining procedures. Suboptimal configurations can therefore lead to instability or performance degradation \cite{li2021differentiable, ding2024pruning}.

In parallel, attribution-based methods aim to quantify the functional contribution of individual attention heads. Techniques such as self-attention attribution \cite{hao2021self} provide fine-grained interpretability by analyzing token-level interactions and gradient-based influence scores. However, these approaches primarily serve as diagnostic tools and rely on manually selected pruning thresholds. Moreover, both optimization-based and attribution-based methods typically evaluate attention heads in isolation, without explicitly accounting for functional redundancy or complementarity among heads.

\subsection{Automation and Interpretability in Pruning} Recent tools have sought to bridge the gap between interpretability and implementation through visual analytics frameworks. The BHPVAS system \cite{liu2024bhpvas} computes importance, stability, and similarity scores to guide manual head pruning via a human-in-the-loop interface. While effective for qualitative analysis, such manual workflows are hard to scale and lack the reproducibility required for fully automated deployment pipelines. Furthermore, these systems require the user to possess domain expertise to interpret the visualizations, leaving the core challenge of autonomous decision-making unsolved.

Our work addresses these limitations by extending the Automatic Complementary Separation Pruning (ACSP) methodology \cite{levin2025automatic} to attention head pruning in Transformer architectures. While ACSP was originally developed for linear and convolutional layers, operating in a layer-wise manner, we adapt its core principles to the Transformer setting by formulating attention head pruning as a global selection problem over the entire network. Rather than treating each layer independently, our approach considers all attention heads jointly, enabling the identification of complementary and {nonredundant} heads across layers. To support this global perspective, we introduce attention-specific signature extraction and padding-aware interpolation mechanisms tailored to the variable-length nature of Transformer inputs. This extension results in a post-hoc pruning pipeline that produces a globally optimized attention head configuration without requiring predefined sparsity targets.

\section{Methodology}


\begin{algorithm}[b!]
\small
\caption{\textsc{\scalebox{0.92}{Complementary Attention Head Pruning}}} \label{alg:acsp_headprune}\KwIn{Transformer model $F(D;W)$, dataset $D = (X,Y)$}
\KwOut{Pruned and optimized model $F^*$}

\assign{\mathcal{H}} collection of all attention heads in $F(D;W)$\;
\assign{N} $|\mathcal{H}|$ \Comment*[r]{total number of heads}

\text{graph\_space} $\gets$ construct graph space over $\mathcal{H}$\;

\assign{w} compute salience weights for $\mathcal{H}$\;

\assign{\Phi} $\emptyset$ \Comment*[r]{MSS array}
\For{$k \gets 2$ \KwTo $N$}{
    Run $k$-Medoids on $\text{graph\_space}$\;
    $\Phi[k] \gets$ Calculate MSS\;
}

\assign{k^*} apply Kneedle algorithm on $\Phi$\;
\assign{\mathcal{H}^*} $\{\arg\max_{h \in \mathcal{C}_i} w(h)\}_{i=1}^{k^*}$\;
\assign{\mathcal{H}^*} keep $\arg\max w(h)$ per empty layer\;
\assign{F} prune all $h \notin \mathcal{H}^*$ from $F(D;W)$\;
\assign{F^*} fine-tune $F$ on $D$\;





\end{algorithm}

The overall architecture of the proposed CAHP framework is summarized in Algorithm~\ref{alg:acsp_headprune}.

\subsection{Notation and Method Overview}
We consider a Transformer-based encoder model $F$ consisting of $L$ layers. Each layer $l \in \{1, \dots, L\}$ contains a Multi-Head Attention (MHA) block with $H$ attention heads. The global set of all attention heads in the model is denoted as $\mathcal{H} = \{h_{1,1}, \dots, h_{L,H}\}$, where the total number of heads is $N = L \times H$. Given a dataset $D = \{(x_i, y_i)\}_{i=1}^M$ with $C$ unique classes, our objective is to identify a subset of heads $\mathcal{H}^* \subseteq \mathcal{H}$ that reduces redundancy while preserving the model’s predictive performance.

The proposed framework extends the Automatic Complementary Separation Pruning (ACSP) methodology \cite{levin2025automatic} by adapting its core principles to the structural characteristics of Transformer architectures. While ACSP was originally developed for iterative, layer-wise pruning in linear and convolutional networks, we generalize this approach to a global attention head selection strategy. This design choice is motivated by the observation that the total number of attention heads in a Transformer is much smaller than the number of components found in dense architectures, rendering localized, layer-level searches insufficient for clustering. Moreover, prior work on Transformer interpretability has shown that functionally specialized attention heads, such as those capturing syntactic or positional information, are distributed unevenly across layers. Treating all attention heads as a global candidate pool therefore enables the framework to identify complementary and functionally significant heads across the network, rather than being constrained by rigid layer boundaries.

\begin{figure*}[t]
\centering
\includegraphics[width=1.2\columnwidth]{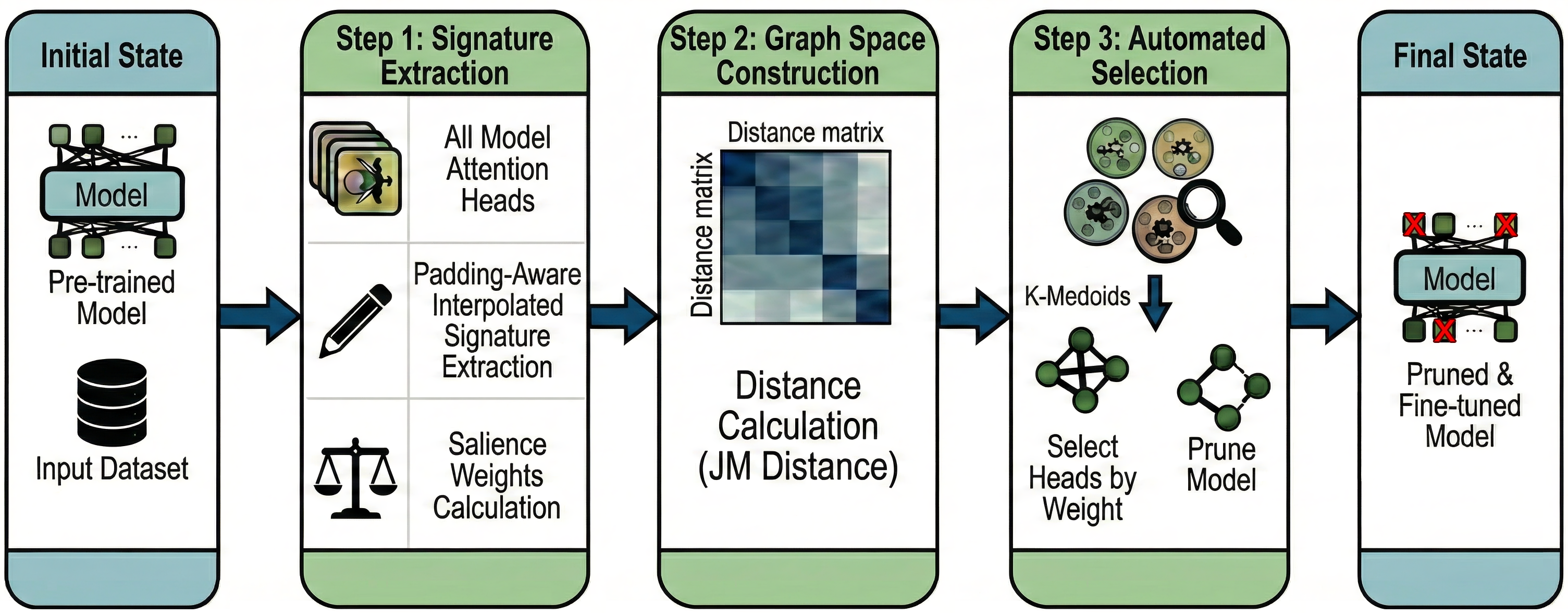}
\caption{Overview of the CAHP pipeline. (i) Signature Extraction: Heads are profiled using padding-aware interpolation and salience weighting. (ii) Graph Space Construction: Pairwise JM distances are mapped to a low-dimensional manifold via t-SNE. (iii) Automated Selection \& Recovery: Optimal clusters are identified via MSS and the Kneedle algorithm, with salience-lead representatives retained for the pruned architecture and subsequent performance recovery.}
\label{fig:cahp_pipeline}
\end{figure*}

Beyond these architectural shifts, our methodology introduces fundamental adaptations to handle the distinct nature of self-attention data. Specifically, we replace post-activation feature maps with attention softmax matrices, {necessitating} a padding-aware interpolation step to normalize variable-length sequences into constant, comparable ``signatures''. Furthermore, because Transformer heads lack the explicit filter magnitudes found in convolutional or linear layers, we substitute static weight metrics with a dynamic gradient-based proxy. These adaptations ensure the framework can effectively categorize and prioritize heads based on their functional contribution, transforming the original layer-wise logic into a robust, global-aware pipeline tailored for natural language processing.

{The resulting CAHP pipeline is illustrated in Fig.~\ref{fig:cahp_pipeline}.} The process consists of three primary stages: (i) signature extraction, where each head's behavior is profiled across the dataset; (ii) graph space construction, where statistical distances between heads are mapped into a low-dimensional manifold; (iii) automated pruning and recovery, utilizing a knee-finding algorithm to determine the optimal architecture followed by fine-tuning to recover performance.

\subsection{Transformer-Specific Signature Extraction}
To quantify the behavior of each head $h \in \mathcal{H}$, we extract a statistical signature derived from its attention maps. In Transformer models, attention maps are input-dependent and vary with the length of the processed sequence, reflecting the dynamic nature of self-attention.  To ensure a consistent and comparable representation across samples, we implement a padding-aware interpolation procedure that normalizes attention maps to a fixed resolution before signature extraction.

For a given input sequence $x_i$, let $A_{l,h} \in \mathbb{R}^{S \times S}$ be the attention weight matrix produced by head $h$ in layer $l$, where $S$ represents the maximum sequence length. To prevent the statistical signatures from being corrupted by padding tokens, we utilize the model's attention mask to crop the matrix to its actual sequence length $S_{real} \times S_{real}$. This cropped matrix is then resized to a fixed spatial resolution $B \times B$ using bilinear interpolation, ensuring that signatures are comparable across variable-length inputs. From these resized matrices, we derive a statistical profile for each head $h$ across each class $c$, characterized by the class-conditional mean $\mu_{h,c}$ and variance $\sigma^2_{h,c}$ of the flattened attention weights.

\subsection{Graph Space Construction and Separation Metrics}
Consistent with the theoretical foundation of the ACSP framework \cite{levin2025automatic}, we quantify head redundancy by measuring the statistical separation of head behaviors across class pairs. {We formulate this as a clustering and representative selection problem, grouping heads with similar behavioral signatures to systematically reduce redundancy while preserving functional diversity.} For each head $h \in \mathcal{H}$ and every possible pair of classes $(c_1, c_2)$, we compute the Jeffries-Matusita (JM) distance. The JM distance between two normal distributions is derived from the Bhattacharyya distance $\mathcal{B}$:

\begin{equation}
\mathcal{B} = \frac{1}{8} \frac{(\mu_1 - \mu_2)^2}{\sigma_1^2 + \sigma_2^2} + \frac{1}{2} \ln \left( \frac{\sigma_1^2 + \sigma_2^2}{2\sigma_1\sigma_2} \right).\label{eq1}
\end{equation}

\begin{equation}
\mathrm{JM} = 2(1 - e^{-\mathcal{B}})\label{eq2}.
\end{equation}

{The JM distance provides a principled and bounded measure of spectral separability, utilizing its saturated behavior to ensure numerical stability during the construction and projection of the high-dimensional feature matrix \cite{wang2018unsupervised, tolpekin2009quantification}.}

We construct a global feature matrix $X \in \mathbb{R}^{N \times (B^2 \cdot P)}$, where $P = \binom{C}{2}$ represents the number of unique class pairs. Each row in $X$ corresponds to a single head, containing the concatenated JM distance vectors across all class pairs. Finally, we utilize t-distributed Stochastic Neighbor Embedding (t-SNE) to project this high-dimensional feature space into a low-dimensional manifold, referred to as the graph space, which facilitates the identification of architectural redundancies.

\subsection{Automated Selection and Volume Optimization}
To identify the optimal subset of attention heads $\mathcal{H}^* \subseteq \mathcal{H}$, we aim to group heads with similar topological behaviors and identify a single, most-informative representative for each identified functional group. Consistent with the selection logic of the original ACSP framework, we achieve this through a two-stage process: {first determining the pruning volume ($k^*$) based on diminishing marginal gains, and second, performing salience-weighted representative selection within each cluster.}


We first evaluate the compactness and diversity of information within the graph space using the Mean Simplified Silhouette (MSS) metric \cite{levin2024gb, levin2025graph,levin2025automatic}. To identify the natural grouping of the model’s attention mechanism, we perform global $k$-medoids clustering iteratively for every possible value $k \in \{2, \dots, N\}$. For each increment of $k$, we calculate the corresponding MSS score, thereby generating a performance curve that represents the trade-off between model complexity and the preservation of functional diversity.

Unlike standard clustering indices that focus on local cluster density, MSS measures the separation between a data point and all other cluster centers. This ensures that a selection of $k$ representatives is not only well-associated with their respective clusters but also well distributed across the graph manifold. By using the Kneedle algorithm \cite{satopaa2011finding} to identify the knee point of the $MSS(k)$ curve, we determine the specific $k^*$ at which the marginal gain in functional representation significantly diminishes, thereby establishing the optimal size for $\mathcal{H}^*$.

For a given clustering of $k$ medoids, the MSS index is calculated by defining the coefficients $a(i)$ and $b(i)$ for every head $i \in \mathcal{H}$. Here, $a(i)$ represents the distance between head $i$ and its assigned cluster medoid $C_h$:

\begin{equation}
a(i) = d(x_i, C_h).\label{eq3}
\end{equation}

The value $b(i)$ denotes the average distance of head $i$ from the medoids of all other clusters $C_l$ where $l \neq h$:

\begin{equation}
b(i) = \text{avg}_{l \neq h} \left( d(x_i, C_l) \right).\label{eq4}
\end{equation}

The MSS score for an individual head is then defined as:
\begin{equation}
\mathrm{MSS}(i) = 1 - \frac{a(i)}{b(i)}\label{eq5}
\end{equation}

and the global MSS index is the average of these scores across all $N$ heads.

Once the optimal volume $k^*$ is established via the Kneedle algorithm, our framework implements a salience-weighted selection strategy within each identified redundancy cluster. While the original ACSP methodology identifies optimal components based on inherent feature weights, our approach utilizes gradient-based head masking as a proxy for head importance. For each cluster $k \in \{1, \dots, k^*\}$, we identify the most significant representative based on the attribution score $w_{l,h}$. These scores are computed through gradient-based head masking, similar to the attribution framework in \cite{liu2024bhpvas}:

\begin{equation}
w_{l,h} = E_x \left| \frac{\partial \mathcal{L}(x)}{\partial d_{l,h}} \right|\label{eq6}
\end{equation}

where $d_{l,h}$ is a multiplicative mask applied to the output of head $h$. {To ensure numerical consistency, these scores are normalized via a standard min-max transformation to the range $[0, 1]$, facilitating a uniform comparison of functional significance regardless of layer depth or gradient variance.} By picking the head with the maximum $w$ within each redundancy cluster, our method ensures that we retain the most functionally significant representative of each behavior group. Consequently, the chosen subset of heads $\mathcal{H}^*$ is guaranteed to complement each other and fully represent the functional diversity of the original architecture.

To maintain architectural integrity {during the removal of the complementary set $\mathcal{H} \setminus \mathcal{H}^*$}, we enforce a layer-safety constraint. While our global strategy identifies a diverse subset $\mathcal{H}^*$, removing a layer in its entirety causes structural failures in standard deep learning frameworks. To avoid modifications to the underlying model architecture, we implement a procedural safeguard: if a layer is fully pruned, we retain the single attention head with the highest salience score $w$ in that layer.


{Finally, targeted fine-tuning is applied to recover performance, completing the transformation from the original dense architecture to a functionally optimized, sparse model.}

\section{Experimental Evaluation}
In this section, we evaluate the performance of the CAHP framework against state-of-the-art head pruning methods. We assess the framework's ability to maintain predictive accuracy across varying compression levels and architectural scales.

\subsection{Experimental Setup}

{We evaluate our framework on SST-5 (Stanford Sentiment Treebank), a five-class sentiment classification task sensitive to subtle semantic variations, and MNLI (Multi-Genre Natural Language Inference), an entailment task. Performance is reported on the SST-5 test set and MNLI dev-mismatched sets, with results averaged over 10 and 3 random seeds, respectively, to ensure robustness.}


{We utilize cased BERT-base ($L=12, H=12$) and BERT-large ($L=24, H=16$) to evaluate performance across different model scales and head capacities. Cased variants are selected because capitalization signals emotional intensity and semantic emphasis -- key indicators in sentiment analysis.}

\subsection{Implementation Details}
To ensure a rigorous and reproducible comparison, we detail the core engineering choices and hyperparameter configurations utilized in our study.

\begin{enumerate}

    \item \textbf{Feature Representation and Weighting:} Before graph construction, attention matrices are processed by removing padding and interpolating to a fixed resolution, using $B=32$ for SST-5 and $B=48$ for MNLI, to ensure topological consistency. These resolutions are selected to correspond approximately to the 70th percentile of tokenized sequence lengths in their respective datasets. Salience weights $w$ are calculated using a 25\% training data calibration subset. To maintain a low memory footprint when processing large-scale datasets, we utilize Welford’s algorithm \cite{knuth2014art} as an online streaming method to compute the per-class mean and variance profiles. These stabilized statistics serve as the primary inputs for calculating the JM distances within the graph space.


    \item \textbf{Automated Selection Logic:} To identify the optimal cluster count $k^*$, we evaluate the MSS curve across the full range of possible components. For each $k$, we partition the graph space using the FasterPAM algorithm \cite{schubert2021fast}, an optimized $k$-medoids variant that ensures efficient convergence. The knee point is located via the Kneedle algorithm \cite{satopaa2011finding}, which utilizes polynomial fitting to identify the point of maximum curvature. We evaluate polynomial degrees $d \in \{2, \dots, 6\}$ across 10 random seeds {for SST-5, and $d \in \{2, 6\}$ across 3 seeds for MNLI,} to ensure the identified $k^*$ remains stable across varying smoothing intensities and initializations.

\item \textbf{Benchmarking and Fair Comparison:} 
We compare CAHP against four state-of-the-art methods: DSP (Joint and Pipelined variants) \cite{li2021differentiable}, PASS \cite{ding2024pruning}, and AttAttr \cite{hao2021self}. To ensure a fair comparison, all baseline methods are constrained to retain the same number of attention heads ($k^*$) selected by CAHP for each seed. Post-hoc methods (CAHP and AttAttr) are applied to task-specific pre-trained models; similarly, for Pipelined DSP, we bypass the initial fine-tuning stage and perform gate training directly on these models. Conversely, Joint DSP and PASS are initialized from standard HuggingFace checkpoints following their original end-to-end training protocols. All baselines are evaluated using the authors’ recommended hyperparameters, with minor dataset-specific adjustments where required, including a custom integration for SST-5.

    \item \textbf{Hardware and Training:} Experiments were conducted on a system equipped with four NVIDIA Quadro RTX 6000 GPUs (24GB each). {We utilize two GPUs for most experiments, while AttAttr uses a single GPU due to framework-specific constraints.} We maintain a fixed batch size of 32 for both the importance scoring and the final fine-tuning phases. 

    \item \textbf{Post-Pruning Fine-tuning:} After the redundant heads are removed, the model undergoes a brief recovery phase to recalibrate the remaining weights. We fine-tune the pruned architecture for 3 epochs using a learning rate of $2 \times 10^{-5}$ and a linear decay scheduler with a 10\% warmup ratio. To ensure stability and prevent overfitting during this phase, we apply a weight decay of 0.01 and a maximum gradient norm of 1.0, utilizing the same batch size as the initial training.
\end{enumerate}

\subsection{Results and Analysis}
The results of the SST-5 evaluation, summarized in Table \ref{tab:sst5_results}, demonstrate the effectiveness of the CAHP framework across both model scales and the full spectrum of pruning intensities.

\begin{table}[t]
\centering
\caption{Comparative {test set accuracy on SST-5 across different Polynomial Degrees ($d$). Subscripts denote the average percentage of attention heads retained. Accuracy values represent the mean across 10 random seeds.}}
\label{tab:sst5_results}
\renewcommand{\arraystretch}{1.2} 
\begin{tabularx}{0.48\textwidth}{l @{\extracolsep{\fill}} ccccc}
\toprule
\multicolumn{6}{l}{\textit{\textbf{BERT-Large} (384 total heads)}} \\
\midrule
\textbf{Method} & \shortstack{\textbf{Poly 2} \\ {\scriptsize 47.0\%}} & \shortstack{\textbf{Poly 3} \\ {\scriptsize 28.4\%}} & \shortstack{\textbf{Poly 4} \\ {\scriptsize 23.3\%}} & \shortstack{\textbf{Poly 5} \\ {\scriptsize 17.8\%}} & \shortstack{\textbf{Poly 6} \\ {\scriptsize 15.3\%}} \\ 
\midrule
CAHP     & \textbf{55.5\%} & \textbf{54.8\%} & \textbf{54.5\%} & \textbf{53.1\%} & \textbf{53.3\%} \\
Pipelined DSP \cite{li2021differentiable} & 52.1\% & 44.6\% & 38.5\% & 32.5\% & 31.1\% \\
Joint DSP \cite{li2021differentiable}    & 54.6\% & 54.6\% & 53.5\% & 53.0\% & 52.2\% \\
AttAttr \cite{hao2021self}        & 51.1\% & 40.3\% & 35.1\% & 33.4\% & 31.0\% \\
PASS \cite{ding2024pruning}           & 54.6\% & 52.2\% & 40.0\% & 35.7\% & 32.2\% \\
\midrule
\midrule
\multicolumn{6}{l}{\textit{\textbf{BERT-Base} (144 total heads)}} \\
\midrule
\textbf{Method} & \shortstack{\textbf{Poly 2} \\ {\scriptsize 47.9\%}} & \shortstack{\textbf{Poly 3} \\ {\scriptsize 29.3\%}} & \shortstack{\textbf{Poly 4} \\ {\scriptsize 25.1\%}} & \shortstack{\textbf{Poly 5} \\ {\scriptsize 19.8\%}} & \shortstack{\textbf{Poly 6} \\ {\scriptsize 18.1\%}} \\ 
\midrule
CAHP     & \textbf{52.7\%} & \textbf{52.1\%} & \textbf{51.9\%} & \textbf{51.1\%} & \textbf{51.0\%} \\
Pipelined DSP \cite{li2021differentiable} & 52.5\% & 47.6\% & 45.0\% & 39.7\% & 38.2\% \\
Joint DSP \cite{li2021differentiable}    & 51.0\% & 50.0\% & 49.5\% & 48.6\% & 48.4\% \\
AttAttr \cite{hao2021self}        & 48.6\% & 42.8\% & 41.3\% & 39.0\% & 38.1\% \\
PASS \cite{ding2024pruning}           & 51.1\% & 47.3\% & 39.8\% & 41.8\% & 38.2\% \\
\bottomrule
\end{tabularx}
\end{table}

Compared to unpruned baselines (BERT-Large: 56.8\%; BERT-Base: 53.6\%), CAHP preserves performance despite significant architectural reductions. In the moderate Poly 2 regime ($\sim$47\% retention), accuracy remains within 1.3\% of the baseline for BERT-Large and 1\% for BERT-Base. Even under extreme pruning (Poly 6), where only $\sim$15.3\% and $\sim$18.1\% of heads remain, accuracy loss is restricted to 3.5\% and 2.6\%, respectively. This confirms that most attention heads in pre-trained Transformers are redundant and that a carefully selected topological subset can sustain predictive capacity.

\begin{figure}[!b]
\centering
\includegraphics[width=0.9\columnwidth]{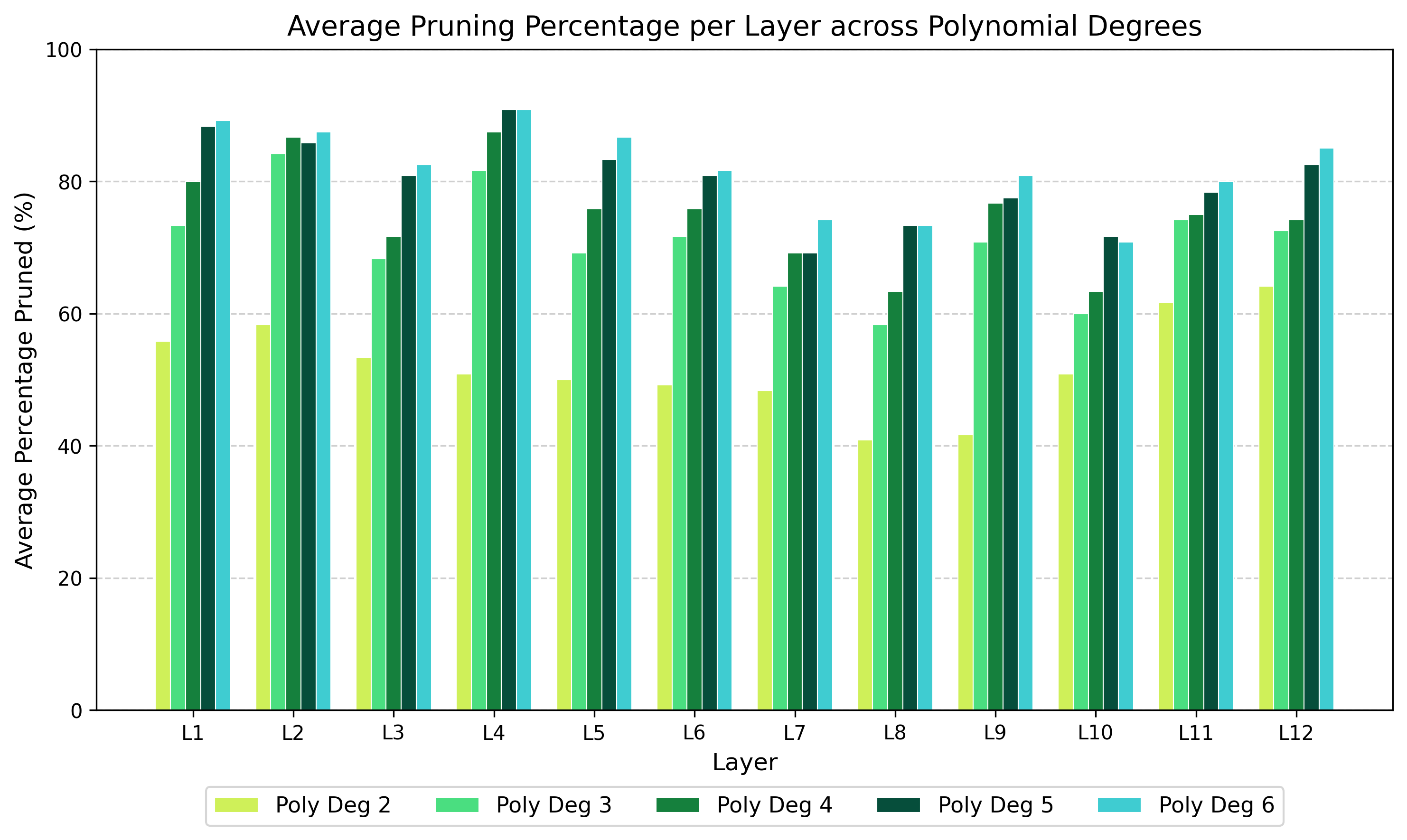}
\caption{Average pruning percentage per layer across various polynomial degrees ($d$) for BERT-Base.}
\label{fig:poly_trends}
\end{figure}

To connect the observed performance trends with architectural behavior, we analyze the layer-wise distribution of retained attention heads across polynomial degrees. As illustrated in Fig.~\ref{fig:poly_trends}, the average pruning ratio varies systematically across model depth, indicating {nonuniform} retention. At lower degrees (e.g., Poly~2), pruning is most pronounced in the early and final layers, whereas intermediate layers (L8–L9) consistently retain a larger fraction of heads. As the degree increases toward Poly~6, the distribution becomes more uniform; however, intermediate layers continue to exhibit lower pruning rates than the periphery. This persistent pattern suggests that our selection mechanism prioritizes intermediate layers, which are empirically less aggressively pruned across all compression levels, thereby preserving the internal transformation stages most critical for performance under high sparsity.

{In direct comparison with state-of-the-art methods, CAHP consistently achieves the highest accuracy across all polynomial degrees and architectures. AttAttr and Pipelined DSP exhibit sharp performance degradation as sparsity increases, often falling below 40\% accuracy. These frameworks prioritize identifying redundant heads without an integrated weight-recovery phase; following original protocols, we evaluate them without post-pruning fine-tuning to maintain consistency with their intended design. In contrast, CAHP's end-to-end pipeline highlights the critical role of post-pruning fine-tuning in recalibrating the model to its structural constraints. Furthermore, by selecting diverse, complementary functional components rather than relying solely on importance scores, CAHP provides a stable foundation for recovery, preventing the loss of unique linguistic features that ranking methods may discard.}

\begin{figure}[!bt]
\centering
\includegraphics[width=0.9\columnwidth]{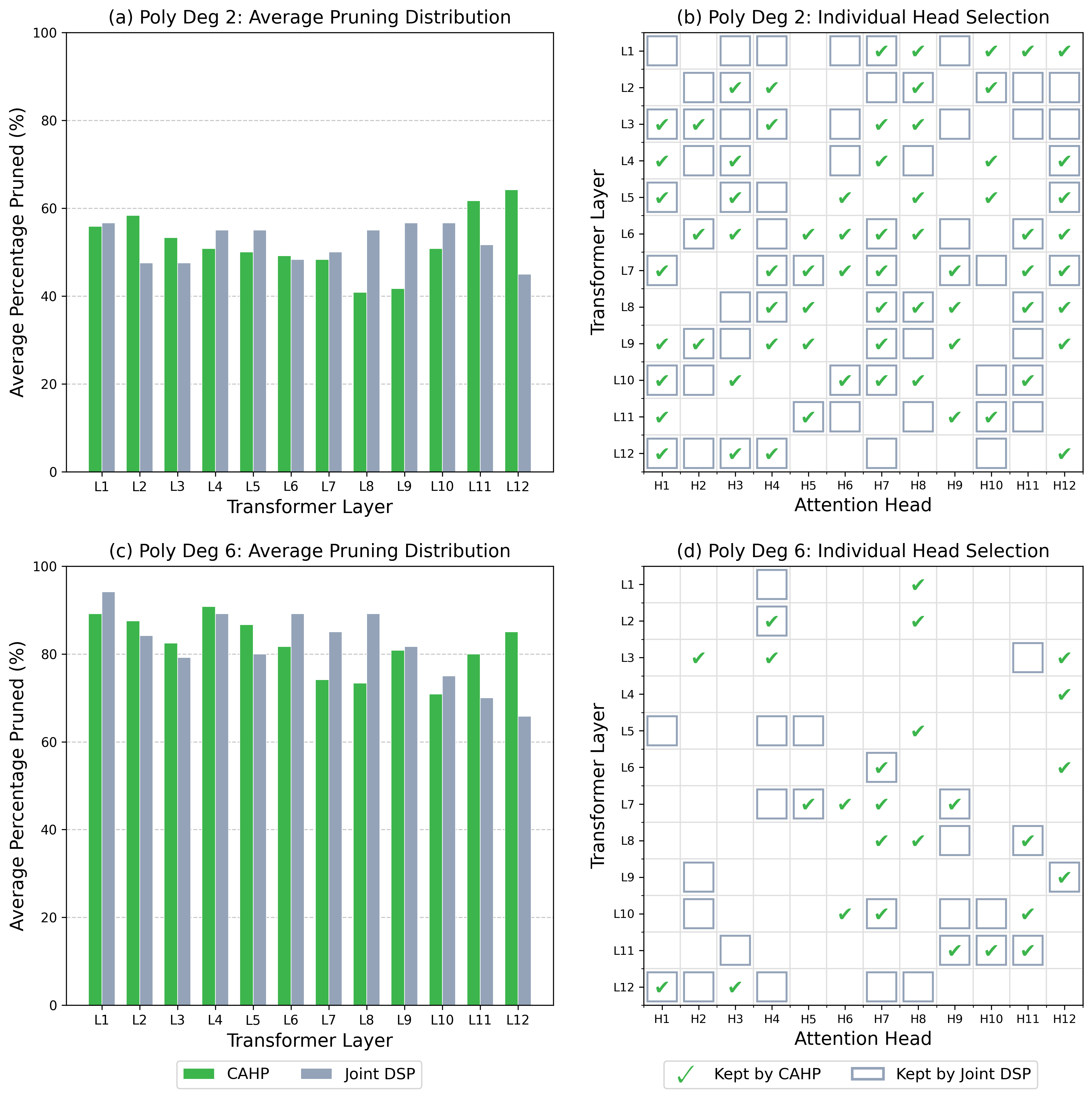}
\caption{Structural comparison of CAHP and Joint DSP pruning for Poly 2 and Poly 6. Plots (a, c) show average pruning ratios across all seeds, while maps (b, d) show head selection for a representative seed, illustrating sub-network emergence at higher complexities.}
\label{fig:pruning_analysis}
\end{figure}

Among baselines, Joint DSP and PASS emerged as the strongest competitors, likely due to co-optimizing weights and masks. However, CAHP outperforms these approaches, particularly in high-sparsity Poly 4–6 regimes. To provide a granular qualitative comparison, we analyze CAHP's selection patterns against Joint DSP, our most competitive baseline, as illustrated in Fig.~\ref{fig:pruning_analysis}. At lower polynomial degrees (Poly~2, top row), while both methods retain comparable total volumes, subfigure (b) reveals distinct architectural decisions; for the representative seed, the methods share only 36 of 69 total heads. The average layer-wise distributions in subfigure (a) confirm this divergence: while Joint DSP exhibits a relatively uniform distribution with a slight bias toward final layers, CAHP demonstrates a strategic preference for the model's interior, shaping the structural core critical at higher sparsity. CAHP's pruning pattern concentrates in deeper layers, consistent with prior work showing mid-to-late attention heads are often redundant \cite{zhang2024finercut, sajjad2023effect, gromov2024unreasonable}.

The difference in strategy becomes pronounced under high sparsity (Poly 6, bottom row). Here, overlap between methods collapses to 11 of 26 heads (subfigure (d)), indicating fundamentally different importance criteria. Pruning ratios in subfigure (c) expose a clear proximity bias in Joint DSP, which concentrates capacity in final layers (L10–L12). This suggests the baseline’s gradient-driven optimization is heavily influenced by proximity to the loss function. In contrast, CAHP retains more heads in intermediate layers (L7–L9), avoiding the unbalanced distributions seen in Joint DSP. By favoring topologically central heads over gradient magnitude, CAHP preserves the intermediate transformation stages critical for representation learning, reducing structural bottlenecks under aggressive compression.

\begin{figure}[!b]
\centering
\includegraphics[width=0.9\columnwidth]{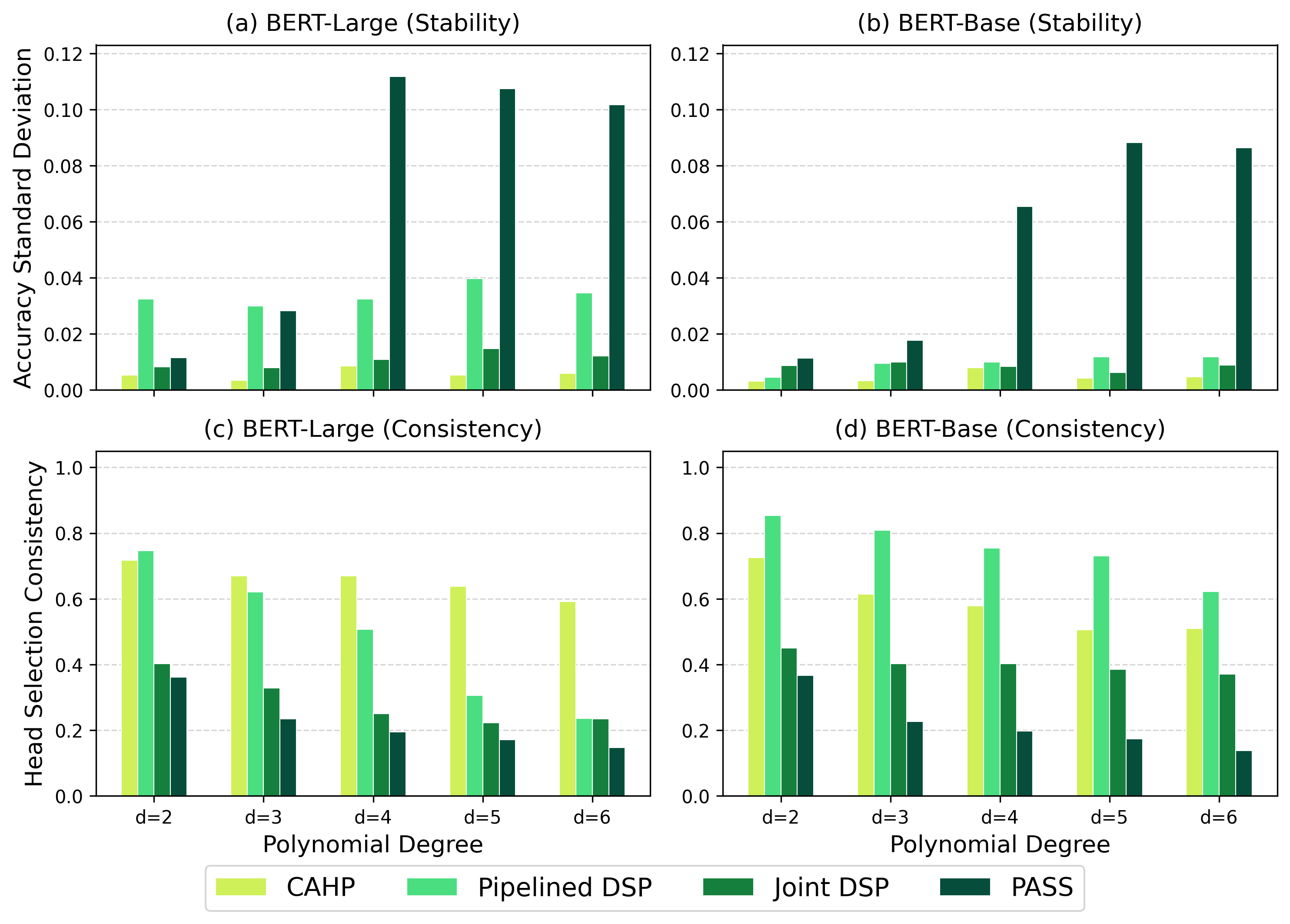}
\caption{Stability and structural consistency on SST-5 across polynomial degrees. Results for BERT-Large {(a, c)} and BERT-Base {(b, d)} demonstrate the impact of pruning regimes on behavior. Panels {(a, b)} show accuracy standard deviation across 10 seeds measuring training stability. Panels {(c, d)} display the average pairwise Jaccard similarity between kept heads across seeds, quantifying the structural consistency of the emergent sparse sub-networks.}
\label{fig:stability_analysis}
\end{figure}

Beyond individual selection patterns, we evaluate the robustness of these architectural decisions across multiple runs. We observed instability in PASS; while some seeds were competitive, others suffered catastrophic drops, leading to lower average accuracies and higher variance. CAHP demonstrated consistent stability across all 10 seeds, suggesting graph-based selection provides a more robust foundation for recovery than stochastic gating. A comparative stability analysis (excluding AttAttr, as its implementation is inherently deterministic) is presented in Fig.~\ref{fig:stability_analysis}. Across all polynomial degrees and frameworks, CAHP exhibits consistently low accuracy variance, indicating stable performance across random seeds. This reliability is further supported by Jaccard similarity analysis: while Pipelined DSP shows high structural overlap primarily in high-retention BERT-Base configurations, CAHP maintains stronger structural consistency under high-compression BERT-Large settings. Even at Poly~6, CAHP preserves substantially higher average pairwise Jaccard similarity than joint-optimization baselines, suggesting that the proposed graph-theoretical selection yields a more stable and functionally grounded head selection under aggressive pruning.

To further validate CAHP's robustness, we evaluate performance on MNLI using the boundary regimes from our SST-5 analysis (Poly 2 and Poly 6). As shown in Table \ref{tab:mnli_results}, CAHP scales effectively to large-scale inference tasks, mirroring our smaller-scale findings. Compared to unpruned baselines (BERT-Large: 86.5\%; BERT-Base: 84.2\%), CAHP exhibits minimal degradation under moderate pruning, with accuracy drops restricted to 0.6\% and 1.2\% for BERT-Large and BERT-Base, respectively. Even under extreme Poly 6 compression (retention $<$ 20\%), performance remains stable with losses of only 2.5\% and 2.4\%, respectively.

\begin{table}[!tb]
\centering
\caption{Comparative accuracy on the MNLI dev-mismatched set across different Polynomial Degrees ($d$). Subscripts denote the average percentage of attention heads retained; accuracy values represent the mean across 3 random seeds.}
\label{tab:mnli_results}
\setlength{\tabcolsep}{2pt} 
\renewcommand{\arraystretch}{1.2}
\begin{tabularx}{\columnwidth}{l *{4}{>{\centering\arraybackslash}X}}
\toprule
 & \multicolumn{2}{c}{\shortstack{\textbf{BERT-Large} \\ (384 total heads)}} & \multicolumn{2}{c}{\shortstack{\textbf{BERT-Base} \\ (144 total heads)}} \\
\cmidrule(lr){2-3} \cmidrule(lr){4-5}
\textbf{Method} & \shortstack{\textbf{Poly 2} \\ {\scriptsize 45.7\%}} & \shortstack{\textbf{Poly 6} \\ {\scriptsize 14.8\%}} & \shortstack{\textbf{Poly 2} \\ {\scriptsize 47.2\%}} & \shortstack{\textbf{Poly 6} \\ {\scriptsize 19.4\%}} \\ 
\midrule
CAHP                                      & \textbf{85.9\%} & 84.0\%          & \textbf{83.0\%} & \textbf{81.8\%} \\
Pipelined DSP \cite{li2021differentiable} & 83.9\%          & 57.8\%          & 81.0\%          & 72.5\%          \\
Joint DSP \cite{li2021differentiable}     & \textbf{85.9\%} & \textbf{84.5\%} & 82.8\%          & 81.2\%          \\
AttAttr \cite{hao2021self}                & 85.4\%          & 48.8\%          & 80.1\%          & 63.5\%          \\
PASS \cite{ding2024pruning}               & 85.4\%          & 83.3\%          & 82.8\%          & 77.2\%          \\
\bottomrule
\end{tabularx}
\end{table}

Compared to baseline frameworks, CAHP remains highly competitive with the strongest joint-optimization methods. On BERT-Large at Poly 2, CAHP and Joint DSP achieve identical accuracies, matching state-of-the-art performance for this pruning level. While Joint DSP maintains a 0.5\% edge over CAHP at the extreme Poly 6 regime for the Large model, the two remain comparable, underscoring the strength of graph-based selection against intensive co-optimization. The performance gap is more pronounced in BERT-Base experiments, where CAHP achieves the highest accuracy across both polynomial degrees. Notably, post-hoc methods without integrated weight recovery, such as AttAttr and Pipelined DSP, collapse under high sparsity on this larger dataset, whereas PASS shows improved stability on MNLI. These observations confirm that CAHP’s ability to identify a functionally grounded topological core sustains performance on complex benchmarks even as architectural capacity is aggressively reduced.

\subsection{Computational Efficiency and Practicality}
Beyond predictive performance, we evaluate the practical feasibility of CAHP by comparing its end-to-end runtime against established baselines (Table III). Across architectures and datasets, CAHP demonstrates a runtime comparable to joint optimization methods such as Joint DSP and PASS. While CAHP introduces an additional graph-based selection phase, the overall pruning pipeline remains within the same order of magnitude as iterative gating approaches.

Importantly, this computational cost is achieved without requiring predefined sparsity targets. Instead, CAHP relies on polynomial fitting of the MSS curve to automatically determine the number of retained heads in a single pass, avoiding the trial-and-error tuning commonly associated with stochastic pruning frameworks. As a result, the method offers a favorable balance between automation and efficiency, making it well suited for practical deployment scenarios.

\begin{table}[!tb]
\centering
\caption{Approximate total runtime comparison across datasets and architectures. Times are reported in hours (h) and minutes (m).}
\label{tab:runtime_comparison}
\setlength{\tabcolsep}{2pt}
\renewcommand{\arraystretch}{1.2} 
\begin{tabularx}{\columnwidth}{l *{4}{>{\centering\arraybackslash}X}}
\toprule
 & \multicolumn{2}{c}{\textbf{BERT-Large}} & \multicolumn{2}{c}{\textbf{BERT-Base}} \\
\cmidrule(lr){2-3} \cmidrule(lr){4-5}
\textbf{Method} & \textbf{SST-5} & \textbf{MNLI} & \textbf{SST-5} & \textbf{MNLI} \\
\midrule
CAHP & 10m & 5h 40m & 3m & 2h 2m \\
Pipelined DSP \cite{li2021differentiable} & 3m & 1h 22m & 1m & 27m \\
Joint DSP \cite{li2021differentiable} & 8m & 5h 21m & 3m & 1h 44m \\
AttAttr \cite{hao2021self} & 30m & 30m & 5m & 6m \\
PASS \cite{ding2024pruning} & 8m & 5h 38m & 3m & 1h 49m \\
\bottomrule
\end{tabularx}
\end{table}

\section{Conclusion}
In this work, we presented CAHP, an automated, post-hoc framework for structured Transformer pruning that identifies an effective allocation of attention heads based on diminishing marginal performance gains. By shifting the pruning paradigm from importance ranking to graph-based complementary selection, we successfully bridged the gap between theoretical prunability and practical deployment. Unlike optimization-based methods that rely on manual sparsity targets, extensive hyperparameter tuning, or expensive retraining, our approach provides a pipeline that automatically determines the layer-wise distribution of attention heads, achieving a favorable trade-off between model compactness and predictive accuracy.

{Evaluation} on the SST-5 and MNLI benchmarks, conducted across different Transformer model scales, a wide range of sparsity levels, and multiple random seeds, demonstrates that CAHP outperforms state-of-the-art {baselines} under high compression. As observed in our structural comparison with Joint DSP, gradient-driven optimization often leads to a proximity bias, whereby attention heads are preserved in layers closer to the output. In contrast, our topological analysis identifies the functional core of the model within the intermediate layers, a pattern consistent across architectures and both single-sentence and sentence-pair tasks. By preserving this layer-wise structural pattern, CAHP avoids collapse and instability observed in {other} frameworks, {performing robustly} even when fewer than 20\% of the original attention heads are retained.





{Future work should extend our evaluation to generative, unsupervised, and multilingual settings, {moving} beyond the current supervised framework. It may also explore alternative distance metrics and clustering methods {in order} to improve efficiency. Finally, integrating complementary selection into training, rather than post-hoc only, could further enhance stability and performance.}


\section*{Acknowledgment}
Gemini AI was used for language and grammar editing.

\bstctlcite{IEEEexample:BSTcontrol}
\bibliographystyle{IEEEtran}
\bibliography{references}

\end{document}